\documentclass[runningheads]{llncs}
\usepackage[T1]{fontenc}

\usepackage[english]{babel}

\usepackage{amsmath}
\usepackage{amsfonts}
\usepackage{graphicx}
\usepackage[colorlinks=true, allcolors=blue]{hyperref}
\usepackage{subfloat}
\usepackage{cleveref}
\usepackage{multirow}
\usepackage{caption}
\usepackage{subcaption}
\usepackage{booktabs}

\usepackage{xcolor, soul}

\newcommand\hc[1]{{\sethlcolor{cyan}{\hl{#1}}}}
\newcommand\ho[1]{{\sethlcolor{orange}{\hl{#1}}}}

\title{Comparison of attention models and post-hoc explanation methods for embryo stage identification: a case study \thanks{Supported by Nantes Excellence Trajectory (NExT)}}

\titlerunning{Attention and explanation methods for embryo stage identification}

\begin{document}

\author{Tristan Gomez\inst{1}\orcidID{0000-0002-0182-4760} \and
Thomas Fr\'eour\inst{2} \and
Harold Mouch\`ere\inst{1}}
\authorrunning{T. Gomez et al.}
%
\institute{Nantes Universit\'e, Centrale Nantes, CNRS, LS2N, F-44000 Nantes, France \\
\email{\{tristan.gomez,harold.mouchere\}@univ-nantes.fr}\\
\and
Nantes University Hospital, Inserm, CRTI, Inserm UMR 1064, F-44000 Nantes, France\\
\email{thomas.freour@chu-nantes.fr}}
\maketitle

\begin{abstract}
An important limitation to the development of AI-based solutions for In Vitro Fertilization (IVF) is the black-box nature of most state-of-the-art models, due to the complexity of deep learning architectures, which raises potential bias and fairness issues.
The need for interpretable AI has risen not only in the IVF field but also in the deep learning community in general.
This has started a trend in literature where authors focus on designing objective metrics to evaluate generic explanation methods.
In this paper, we study the behavior of recently proposed objective faithfulness metrics applied to the problem of embryo stage identification.
We benchmark attention models and post-hoc methods using metrics and further show empirically that (1) the metrics produce low overall agreement on the model ranking and (2) depending on the metric approach, either post-hoc methods or attention models are favored.
We conclude with general remarks about the difficulty of defining faithfulness and the necessity of understanding its relationship with the type of approach that is favored.
\end{abstract}

\section{Introduction}

Infertility is a global health issue worldwide \cite{infertilityAroundTheGlobe}.
One of the most common treatments for infertile couples is In Vitro Fertilization (IVF).
This procedure notably consists of embryo culture for 2-6 days under controlled environmental conditions, leading to intrauterine transfer or freezing of embryos identified as having a good implantation potential by embryologists.
To allow continuous monitoring of embryo development, Time-lapse imaging incubators (TLI) were first released in the IVF market around 2010. 
This time-lapse technology provides a dynamic overview of embryonic in vitro development by taking photographs of each embryo at regular intervals throughout its development. 
TLI appears to be the most promising solution to improve embryo quality assessment methods, and subsequently, the clinical efficiency of IVF \cite{timeLapseCultureWithMorpho}. 
In particular, the unprecedented high volume of high-quality images produced by TLI systems has already been leveraged using deep learning (DL) methods. 
Previous work has notably focused on designing models that automatically identify the embryo development stages, a crucial task to identify embryos with low live-birth potential \cite{cellCount,BlastCellCount}.
However, an important limitation to the development of AI-based solutions for IVF is the black-box nature of most state-of-the-art models, due to the complexity of deep learning architectures, which raises potential bias and fairness issues \cite{interpretableIVF}.
The need for interpretable AI has risen not only in the IVF field but also in the deep learning community in general.
This has started a trend in literature where authors confront users with models' decisions along with various explanations to determine which fits better the users' needs on a particular application \cite{evalSalMap,evalVis,evalXAI}. 
However, the financial cost and the difficulty of establishing a correct protocol make this approach difficult.
Because of these issues, another trend focuses on designing objective metrics to evaluate generic explanation methods \cite{rise,liftcam,gradcampp,metricPaper}.
In this paper, we follow this trend and study the behavior of objective faithfulness metrics recently proposed applied to the problem of embryo stage identification.
We apply faithfulness metrics to attention models and further show empirically that (1) the metrics produce low overall agreement on the model ranking and (2) depending on the metric approach, either post-hoc methods or attention models are favored.

First, we describe the attention models, post-hoc methods, and faithfulness metrics we use in this work.
Secondly, we compute an extensive benchmark and study the behavior of the metrics on 9 saliency approaches, including 5 post-hoc methods and 4 attention models.
We conclude with general remarks about the difficulty of defining faithfulness and the necessity of understanding its relationship with the type of approach that is favored.

\section{Method}

\subsection{Saliency map generation approaches}

We now describe the models and methods used to generate saliency maps.
First, we list several attention models as they integrate the computation of a saliency map (called an attention map) that is used to guide the decision process. 
Secondly, we list generic post-hoc explanation methods that can generate saliency maps for a wide range of models and architectures without requiring model training.

\paragraph{Attention models.}
In this category, we include models featuring a spatial attention layer in their inference phase. 
Note that we do not include visual transformers \cite{vit} are they are difficult to interpret due to their non-local heads, as opposed to the local CNN kernels.
Hu et al. \cite{seeBetterBeforeLooking} proposed a variant of the model originally proposed by Lin et al. \cite{bcnn} with a convolutional module that processes the feature maps and outputs the attention map.
In the rest of this paper, this model is designated as the Bilinear-CNN (B-CNN).
Fukui et al. \cite{ABN} proposed another variant of Lin et al.'s architecture called Attention Branch Network (ABN) where the attention module is also trained to correctly predict the object class along with the regular classification head.
This was applied to the problem of embryo quality prediction \cite{embryoAtt}, a problem related to the one studied here.

Another line of work proposes to use prototypes to increase interpretability. 
Notably, Interpretability By Parts (IBP) \cite{IBP} is a model that improves upon Chen et al.'s work \cite{ProtoPNet} by encouraging the model to generate prototypes presence probabilities close to 0 or 1, to generate more accurate attention maps and improve interpretability.

The last attention model included in this study is called Bilinear Representative Non-Parametric Attention (BR-NPA) \cite{brnpa} and proposes to generate attention maps to guide the model spatially without any dedicated parameter, contrary to the models mentioned above which feature either a convolution attention module or parametric prototypes. 

\paragraph{Post-hoc explanation methods.}
This category proposes generic methods that can be applied to any CNN with feature maps and does not require training the model, contrarily to the attention modules mentioned above. 
Class Activation Map (CAM) \cite{cam} was proposed to visualize the areas that contributed the most to the prediction of one specific class.
Grad-CAM \cite{gradcam} improves upon CAM and proposes to compute the gradients of the feature maps relatively to the class score to identify the maps that contributed the most to the decision.
A weighted average of the maps is then computed to obtain the saliency map.
The first method included here is Grad-CAM++ \cite{gradcampp}, which further replaces feature-wise gradients with pixel-wise positive partial derivatives to compute the saliency map.
We also use Score-CAM \cite{score_cam} and Ablation-CAM \cite{ablation_cam} later proposed to remove gradients by evaluating each feature map's importance by masking the input image according to the activated areas of the map  or by ablation analysis.

We also apply  RISE \cite{rise}, a method that uses random masks to find the areas that impact the class score the most.
The last method is a baseline that we call Activation Map (AM) which consists to visualize the average activation map by computing the mean of the last layer's feature maps.

\subsection{Faithfulness metrics}

The input image is a 3D tensor $I \in \mathbb{R}^{H\times W \times 3}$ and the saliency map is a 2D matrix $S\in \mathbb{R}^{H'\times W'}$ with a lower resolution, $H'<H$ and $W'<W$.
We study the following faithfulness metrics. 

\paragraph{Increase In Confidence (ICC) \cite{gradcampp}.} 

The IIC metrics measures how often the confidence of the model in the predicted class increases when highlighting the salient areas.
First, the input image is masked with the explanation map as follows :

\begin{equation}
    I_m = \textrm{norm}(\textrm{upsamp}(S)) \bullet I,
\end{equation}
where $\textrm{norm}(S)$ is the min-max normalization function, defined as $\textrm{norm}(S) = \frac{S-min(S)}{max(S)-min(S)}$, $\textrm{upsamp}(S)$ is a function that upsample S to the resolution of I, and $\bullet$ is the element-wise product. 
The IIC metric is defined as: 

\begin{equation} 
\textrm{IIC} = \mathbf{1}_{[c_I < c_{I_m}]},
\end{equation}
where $c_I$ is the score of the predicted class with I as input and $c_{I_m}$ is the score of the same class with $I_m$ as input. 
The intuition is that a good saliency map S highlights areas such that when the non-salient areas are removed, the class score increases.
Therefore, maximizing this metric corresponds to an improvement.
Note that this metric is a binary value and is only useful when computing its mean value over a large number of images, as in \cref{results}.

\paragraph{Average Drop (AD) \cite{gradcampp}.}
Similar to IIC, this metric measures the average score drop when highlighting the salient areas.
Using the same masking of the input image, the AD metric computes the relative score difference between the two images I and $I_m$: 

\begin{equation}
AD = \frac{max(0,c_I-c_{I_m})}{c_I}
\end{equation}

Given that when highlighting the salient areas the class score is not supposed to decrease, minimizing this metric corresponds to an improvement.
Note that the metric name implies computing an average value over several images, as is done in \cref{results}.
However, we did not include the averaging operator in the AD definition to keep notation simple and uniform throughout this section.

\paragraph{Average Drop in Deletion (ADD) \cite{liftcam}.}
Jung et al. proposed a variant of the preceding metric which consists to mask the salient areas instead of the non-salient areas.
To do this, the image is masked with the inverse of the saliency map, which highlights the non-salient areas : 

\begin{equation}
    I_{1-m} = (1 - \textrm{norm}(\textrm{upsamp}(S))) \bullet I
\end{equation}

The ADD metric is then defined as follows: 

\begin{equation}
    ADD = \frac{max(0,c_I- c_{I_{1-m}})}{c_I},
\end{equation}

where $c_{I_{1-m}}$ is the score with $I_{1-m}$. 
Contrarily to AD, this metric removes the salient areas and the class score is expected to decrease, which means that maximizing this metric results in an improvement.
Also, as for AD, we dropped the averaging operator to keep notation simple.

\paragraph{Deletion Area Under Curve (DAUC) \cite{rise}.} 
This metric evaluates the reliability of the saliency maps by progressively masking the image starting with the most important areas according to the saliency map and finishing with the least important.
First, $S$ is sorted and parsed from the highest element to its lowest element.
At each element $S_{i'j'}$, we mask the corresponding area of I by multiplying it by a mask $M^k \in \mathbb{R}^{H\times W}$, where 

\begin{equation}
    M_{ij}^k= 
\begin{cases}
    0,& \text{if } i'r<i<i'(r+1)~\text{and}~j'r<j<j'(r+1)\\
    1,              & \text{otherwise,}
\end{cases}
\end{equation}
and $r= H/H' = W/W'$.
After each masking operation, the model $m$ runs an inference with the updated version of I, and the score of the initially predicted class  is updated, producing a new score $c_k$: 
\begin{equation}
     c_k = m(I \cdot \prod_{\tilde{k}=1}^{\tilde{k}=k} M^{\tilde{k}} ),
\end{equation}
where $k\in \{1,...,H' \times W'\}$.
Secondly, once the whole image has been masked, the scores $c_k$ are normalized by dividing them by the maximum $\underset{k}{max}~c_k$ and then plotted as a function of the proportion $p_k$ of the image that is masked.
The DAUC is finally obtained by computing the area under the curve (AUC) of this graph.
The intuition behind this is that if a saliency map highlights the areas that are relevant to the decision, masking them will quickly result in a large decrease in the initially predicted class score, which in turn will minimize the AUC.
Therefore, minimizing this metric corresponds to an improvement.

\paragraph{Insertion Area Under Curve (IAUC) \cite{rise}.} 
Instead of progressively masking the image, the IAUC metric starts from a blurred image and then progressively unblurs it by starting from the most important areas according to the saliency map.
Similarly, if the areas highlighted by the map are relevant for predicting the correct category, the score of the corresponding class (obtained using the partially unblurred image) is supposed to increase rapidly.
Maximizing this metric corresponds to an improvement.

\paragraph{Deletion Correlation (DC) \cite{metricPaper}.}
The DC metric also consists of gradually masking the input image by following the order suggested by the saliency map, but instead of computing the area under the class score/pixel rank curve, it computes the linear correlation of the class score variations and the saliency scores.
Once the scores $c_k$ have been computed, we compute the variation of the scores $v_k = c_k - c_{k+1}$.
The DC metric is obtained by the linear correlation coefficient between the $v_k$ and the $s_k$ where $s_k$ is the saliency score of the area masked at step k.
As it represents the correlation between the saliency of a pixel and its impact on the class score, maximizing this metric is an improvement.

\paragraph{Insertion Correlation (IC) \cite{metricPaper}.}
Similarly, the IC metric is inspired by IAUC and starts from a  blurred image, and gradually reveals the image according to the saliency map. 
Once the image is totally revealed, the score variations are computed $v_k = c_{k+1}-c_k$ and the linear correlation of the $v_k$ with the $s_k$ is computed.
This correlation metric should also be maximized to correspond to an improvement.

\subsection{The embryo development stage dataset}

The images used in this work are microscopic images of growing embryos between their first and fifth day of development in vitro.
During these five days, the embryo grows through 16 successive development stages that we note sPB2, sPNa, sPNf, s2, s3, s4, s5, s6, s7, s8, s9+, sM, sSB, sB, sEB, and sHB.
These stages are delimited by 16 events called tPB2, tPNa, tPNf, t2, t3, t4, t5, t6, t7, t8, t9+, tM, tSB, tB, tEB, tHB and defined by Ciray et al. \cite{phasesDefinition}.
The timings of each event are valuable information to determine if the quality of the embryo is high enough to be transferred to the uterus \cite{timeLapseCultureWithMorpho}. 

The raw dataset consists of 704 videos that we split equally into one training/validation set and one test set. 
From each set and each video, we extract one-third of the frames regularly spaced  the video.
The images are in grayscale with a $500\times500$ resolution.
The total size of each set is 29843 images for the training/validation set and 28282 images for the test set. 

\begin{figure}
    \centering
    \begin{subfigure}[t]{0.45\textwidth}
        \includegraphics[width=\textwidth]{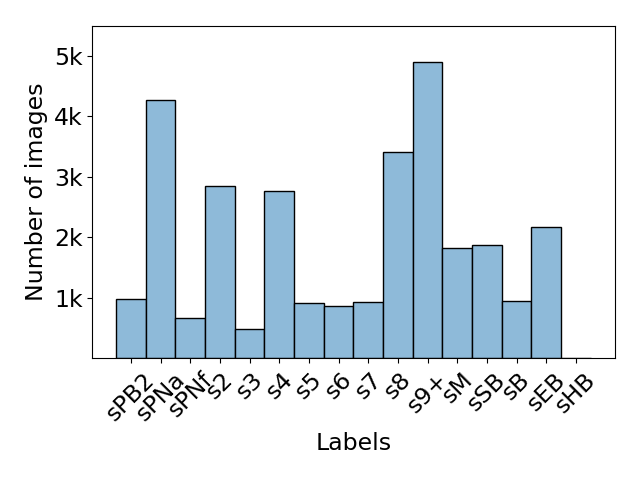}\caption{Training/validation set}
    \end{subfigure}
    \begin{subfigure}[t]{0.45\textwidth}
        \includegraphics[width=\textwidth]{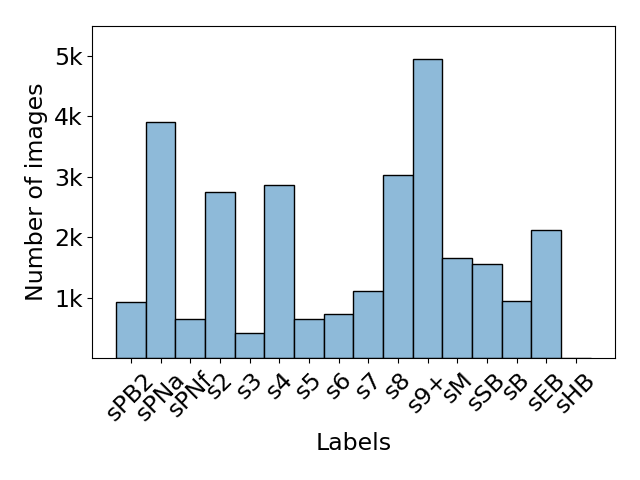}\caption{Test set}
    \end{subfigure}
    \caption{The class distribution of the dataset used. There are only 5 and 15 sHB samples respectively in the training/validation set and the test set as this stage usually occurs after the video recording has been stopped. \label{classDistr}}
\end{figure}

\begin{figure}[ht!b]
    \resizebox{\textwidth}{!}{
    \begin{tabular}{cccccccc}
    \includegraphics[width=0.110\textwidth,trim={0.5cm 0.5cm 0.5cm 0.5cm},clip]{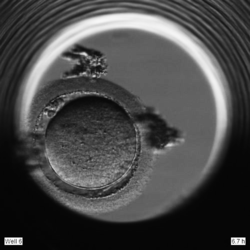} & 
    \includegraphics[width=0.110\textwidth,trim={0.5cm 0.5cm 0.5cm 0.5cm},clip]{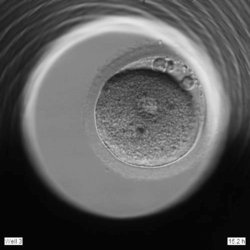} & 
    \includegraphics[width=0.110\textwidth,trim={0.5cm 0.5cm 0.5cm 0.5cm},clip]{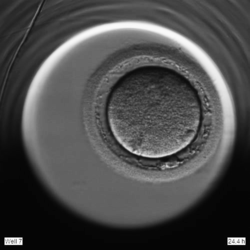} & 
    \includegraphics[width=0.110\textwidth,trim={0.5cm 0.5cm 0.5cm 0.5cm},clip]{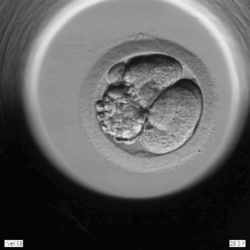} &
    \includegraphics[width=0.110\textwidth,trim={0.5cm 0.5cm 0.5cm 0.5cm},clip]{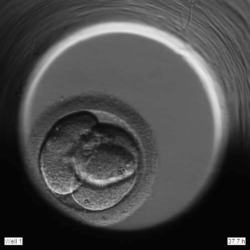} &
    \includegraphics[width=0.110\textwidth,trim={0.5cm 0.5cm 0.5cm 0.5cm},clip]{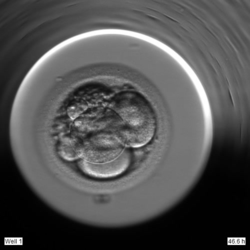} &
    \includegraphics[width=0.110\textwidth,trim={0.5cm 0.5cm 0.5cm 0.5cm},clip]{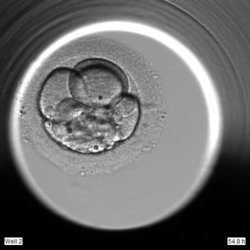} &
    \includegraphics[width=0.110\textwidth,trim={0.5cm 0.5cm 0.5cm 0.5cm},clip]{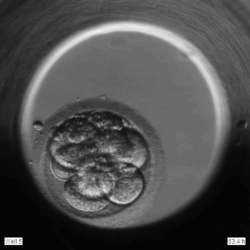} \\
    
    sPB2 & sPNa & sPNf & s2 & s3 & s4 & s5 & s6 \\
    Second polar & Pro-nuclei & Pro-nuclei & \multirow{2}{*}{$2$ cells} &\multirow{2}{*}{$3$ cells} & \multirow{2}{*}{$4$ cells} & \multirow{2}{*}{$5$ cells} & \multirow{2}{*}{$6$ cells} \\
    body detached  & appearance & disappearance & \\
    \includegraphics[width=0.110\textwidth,trim={0.5cm 0.5cm 0.5cm 0.5cm},clip]{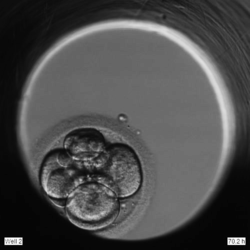} &
    \includegraphics[width=0.110\textwidth,trim={0.5cm 0.5cm 0.5cm 0.5cm},clip]{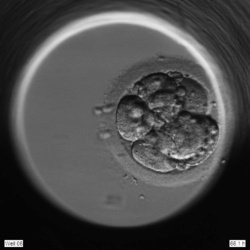} &
    \includegraphics[width=0.110\textwidth,trim={0.5cm 0.5cm 0.5cm 0.5cm},clip]{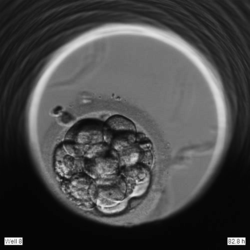} &
    \includegraphics[width=0.110\textwidth,trim={0.5cm 0.5cm 0.5cm 0.5cm},clip]{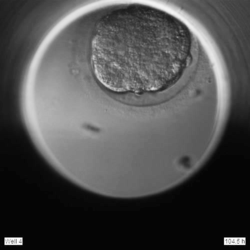} &
    \includegraphics[width=0.110\textwidth,trim={0.5cm 0.5cm 0.5cm 0.5cm},clip]{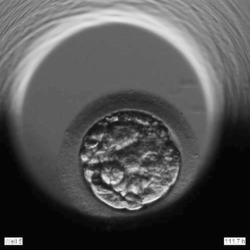} &
    \includegraphics[width=0.110\textwidth,trim={0.5cm 0.5cm 0.5cm 0.5cm},clip]{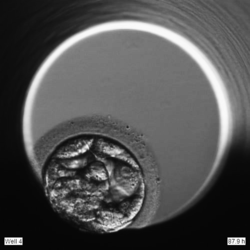} &
    \includegraphics[width=0.110\textwidth,trim={0.5cm 0.5cm 0.5cm 0.5cm},clip]{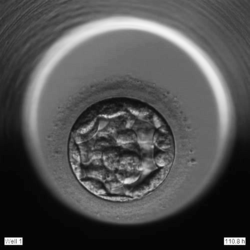} &
    \includegraphics[width=0.110\textwidth,trim={0.5cm 0.5cm 0.5cm 0.5cm},clip]{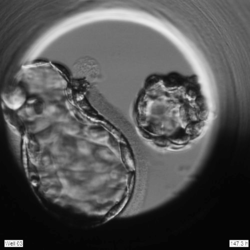} \\
    s7 & s8 & s9+ & sM & sSB & sB & sEB & sHB  \\ 
     \multirow{2}{*}{$7$ cells} & \multirow{2}{*}{$8$ cells} & $9$ cells & End of &   Start of   & Full       & Expanded   & Hatched  \\
    & & or more   & compaction &  blastulation& blastocyst & blastocyst & blastocyst  \\
    \end{tabular}
    }
    \caption{Illustrations of the $16$ development stages used.\label{samples}}
    \end{figure}

We choose to model this problem as an image classification task.
The models studied in this paper are trained to process an embryo image and infer which development stage the embryo is at.
Note that the video recording starts at the tPB2 event which means that the first frames show the sPB2 stage.
The class distribution and some samples are shown respectively in \cref{classDistr} and \cref{samples}.

\section{Implementation details}

The backbone used for all networks is ResNet-50 \cite{resnet}.
Images are augmented during training using a random $448\times448$ crop and random horizontal flipping.
During the test, we extract a center crop of size $448\times448$.
We use 10\% of the training set images for validation.
The models are trained using regular using cross-entropy during 10 epochs and the best model on the validation set is restored for the test phase.
The following hyperparameters were searched on the validation set using the Optuna python framework (\cite{optuna}) with the default sampler (a Tree-structured Parzen Estimator algorithm): the learning rate, momentum, optimizer, batch size, dropout on the classification layer, and weight decay.
We use Pytorch 1.10.2 \cite{torch} and two P100 GPUs.
Following the original work of Petsiuk et al. \cite{rise}, we sampled 4000 masks at a $7\times7$ resolution for the RISE post-hoc method.

\section{Results \label{results}}

In \cref{res} one can see the faithfulness evaluation of the explanation methods and the attention models on the test set.
Given the size of the test set and the computation time of the metrics/approaches, we sample randomly 100 images from the test set and compute the average performance of each approach to each metric on these images.
First, note that we also include the average accuracy per video of the trained models over the whole test set to highlight that all models have similar accuracies.
This means that the differences in faithfulness observed can not be explained by different levels of accuracy but only by the approach (attention model/post-hoc method) or the metrics.
Secondly, the faithfulness metrics do not agree on which model is the best.
For example, depending on the metric, the most faithful explanation method is BR-NPA, RISE, Ablation-CAM, ABN, or Score-CAM and the least faithful is ABN, InterByParts, RISE, BR-NPA, or AM.

\begin{table}[tbh!p]  
    \centering
    \resizebox{\textwidth}{!}{%
    \begin{tabular}{cc|cc|cc|ccc||c}\toprule
Model&Viz. Method&DAUC&IAUC&DC&IC&IIC&AD&ADD&Accuracy\\ 
\multirow{0}{*}{CNN}&AM&$0.134$&$0.308$&$-0.174$&$0.075$&$0.19$&$0.397$&$0.325$&\multirow{5}{*}{$71.0$}\\ 
&Grad-CAM++&$0.1162$&$0.333$&$-0.101$&$0.032$&$0.52$&$0.14$&$0.383$&\\ 
&RISE&$0.113$&$\mathbf{0.457 }$&$-0.221$&$-0.077$&$0.5$&$0.137$&$0.436$&\\ 
&Score-CAM&$0.1079$&$0.315$&$-0.123$&$0.081$&$0.52$&$\mathbf{0.108 }$&$0.362$&\\ 
&Ablation-CAM&$0.0954$&$0.329$&$\mathbf{0.272 }$&$-0.071$&$\mathbf{0.57 }$&$0.111$&$0.328$&\\ 
\hline 
ABN&-&$0.1464$&$0.249$&$-0.186$&$\mathbf{0.136 }$&$0.12$&$0.591$&$0.475$&$71.0$\\ 
InterByParts&-&$0.0876$&$0.115$&$0.196$&$-0.255$&$0.08$&$0.901$&$0.879$&$\mathbf{71.3 }$\\ 
B-CNN&-&$0.0772$&$0.221$&$-0.208$&$0.124$&$0.13$&$0.491$&$0.482$&$70.2$\\ 
BR-NPA&-&$\mathbf{0.0709 }$&$0.261$&$0.185$&$-0.146$&$0.04$&$0.91$&$\mathbf{0.887 }$&$70.7$\\ 
\bottomrule
    \end{tabular}
    }
    \caption{Performance of the studied models and methods.
    \label{res}} 
\end{table}

\begin{figure}[tbh!p]
    \centering
    \includegraphics[width=0.6\textwidth]{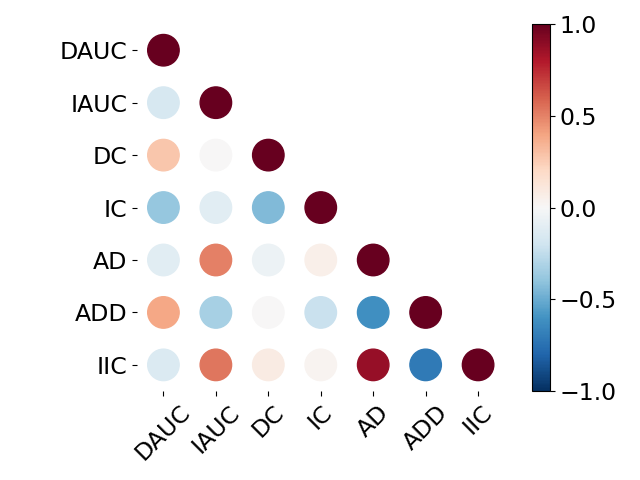}
    \caption{Kendall-tau correlation coefficient between the metrics.
    \label{tau}}
\end{figure}

We propose to use the Kendall-tau correlation coefficient \cite{kendall} to obtain a better idea of the disagreement between the metrics.
This correlation coefficient indicates if two variables have a monotonic relationship with each other.
When close from 1/-1, the variables have an increasing/decreasing relationship, and when close to 0, they are independent.
We represent each metric using the 9 mean performances they have attributed to each approach.
Let $x$ and $y$ be two such lists of performance for two metrics. 
We define a concordant pair, which is a pair of approaches A and B that are ranked the same in $x$ and $y$, i.e. metrics $x$ and $y$ agree that A is better or worse than B.
We also define a discordant pair, where $x$ and $y$ disagree on the ranking of A and B.
Furthermore, we say that a pair is a tie according to $x$ if they have the same ranking in $x$.
Finally, the Kendall-tau correlation coefficient between two metrics $x$ and $y$ is computed as follows:

\begin{equation}
\tau(x,y) = \frac{(P-Q)}{\sqrt{P+Q+U}*\sqrt{P+Q+T}},
\end{equation}
where P is the number of concordant pairs, Q the number of discordant pairs, T the number of ties only in $x$, and U the number of ties only in $y$.
If a tie occurs for the same pair in both $x$ and $y$, it is not added to either T or U.
The coefficient values obtained between each pair of metrics are shown in \cref{tau}. 
Overall, there is a low correlation value between metrics.
This means that metrics globally disagree on the model ranking, with many metrics producing independent ranking (DC and IC have a close from 0 correlation with AD, ADD, IIC, and IAUC) and some metrics even ranking models in reversed order (AD and ADD, ADD and IIC, IC, and DC).
However, some metrics offer similar rankings like AD with IIC or IAUC.

To obtain a clear representation of metrics' similarities with each other, we propose to use a dimension reduction algorithm like t-SNE \cite{tsne}. 
We represented the 9 values given by each of the metrics on each of the 100 images using the dimension reduction algorithm t-SNE.
Note that we exclude the IIC metric as it is a binary metric and will not provide a meaningful ranking on a single image. 
We obtain 6 point clouds in a 9-dimensional space, where each cloud is composed of 100 points and represents one metric, for a total of 600 points. 
We use t-SNE to reduce the dimension of the 600 points at the same time and visualize the result on a single 2D plot.
Note that an appropriate distance function has to be chosen, given that each point represents a ranking. 
Instead of the default Euclidean distance, we use the following distance function based on Kendall's tau correlation: 

\begin{equation}
    D(x,y) = -\log(\frac{1}{2}(\tau(x,y)+1)),
\end{equation}
where $\tau(\cdot,\cdot)$ is the Kendall's tau correlation.
With this metric, we ensure a distance that goes from $0$ to $+\infty$ when the correlation $\tau(x,y)$ goes from 1 to -1.
Furthermore, when $\tau=-1$ we replace its value by $-0.999$ to prevent any numerical errors.
\footnote{Note that we did not use the more recent UMAP algorithm \cite{umap} because the only UMAP implementation available requires the custom distance function to be compiled with Numba, which is currently not possible with the Kendall's tau implementation of the Scikit-Learn python package \cite{scikit-learn}.}.

\begin{figure}[tbh!p]
    \centering
    \includegraphics[width=0.6\textwidth]{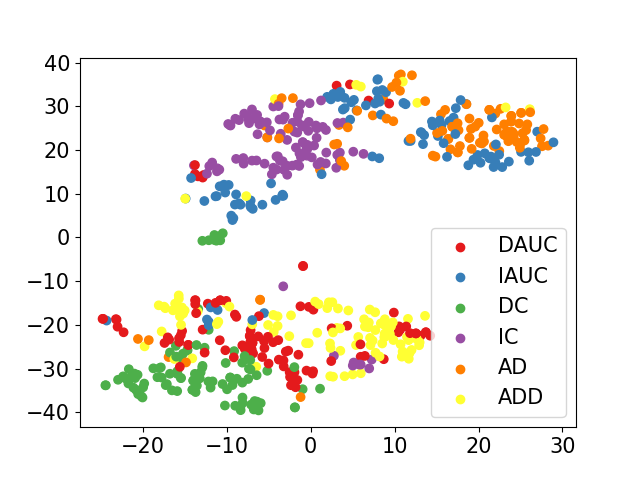}
    \caption{t-SNE projection of the ranking given by metrics on the 100 images. The distance between points is computed using a metric based on Kendall's tau correlation.
    \label{dimred}}
\end{figure}

The plot in \cref{dimred} shows that the metrics are arranged in two groups: \{DAUC, DC, ADD\}, \{IAUC, IC, AD\}.
This structure is not surprising given that DAUC, DC and ADD all consist to mask the important areas of the image, measure the score drop, and favor the model with the lowest class score once the mask(s) are applied. 
Similarly, IAUC, IC, and AD all highlight/reveal the important areas and favors models for which the class score is the highest possible once the important areas have been revealed.
This is why we call 'Mask' and 'Highlight' the two groups of metrics obtained: \{DAUC, DC, ADD\} and \{IAUC, IC, AD, IIC\}.
Note that we also include IIC in the Highlight group as it also consists to highlight the salient areas of the image and expects a score increase as a consequence.
This demonstrates that the method used to measure the impact of an area (Highlight or Mask) of the image majorly impacts the ranking produced.

We propose now to identify the best approaches according to each group of metrics.
We compute the average rank of an approach A on the Highlight and Mask groups as follows: 
\begin{equation}
\begin{split}
r_{Mask}(A) &= \frac{r_{DC}(A)+r_{DAUC}(A)+r_{ADD}(A)}{3}  \\
r_{Highlight}(A) &= \frac{r_{IC}(A)+r_{IAUC}(A)+r_{AD}(A)+r_{IIC}(A)}{4},
\end{split}
\end{equation}
where $r_{X}(A)$ is the rank of A according to the metric X.

\Cref{avg_rank} shows the average ranks computed. 
One can see a link between metric type and approach type where the Mask group seems to favor the Attention models whereas the Highlight group favors the generic post-hoc methods.

\begin{table}[tbh!p]  
    \centering
    \begin{tabular}{ccc|ccc}
    \toprule 
    \multicolumn{3}{c}{Mask} & \multicolumn{3}{c}{Highlight} \\ 
    Average rank & Metric & Type & Average rank & Metric & Type \\ 
    \midrule 
    1.67& \ho{BR-NPA} & Attention &  2.75 &\hc{Score-CAM}&Post-Hoc \\
    2.33& \ho{InterByParts}&Attention &  3.0& \hc{Ablation-CAM}&Post-Hoc \\
    4.33& \ho{B-CNN} &Attention&  3.25 &\hc{Grad-CAM++}& Post-Hoc\\ 
    4.33& \hc{Ablation-CAM} &Post-Hoc& 3.75 &\hc{RISE} &Post-Hoc\\
    5.67& \hc{Grad-CAM++} &Post-Hoc&  4.75 &\hc{AM}& Post-Hoc\\
    5.67& \hc{Score-CAM} &Post-Hoc&  5.5 &\ho{ABN}&Attention \\
    6.67& \ho{ABN} &Attention&  5.5 &\ho{B-CNN}&Attention\\
    6.67& \hc{RISE} &Post-Hoc& 8.0 &\ho{BR-NPA}& Attention\\
    7.67& \hc{AM} &Post-Hoc&   8.5 &\ho{InterByParts}&Attention \\
    \bottomrule
    \end{tabular}
    \caption{Average ranking of the methods according to the two metrics groups: Mask and Highlight. Attention models and post-hoc methods are respectively highlighted in orange and cyan.
    \label{avg_rank}} 
\end{table}

At this point, we identify the most faithful approaches according to each metric group: BR-NPA and Score-CAM.
Next, we propose to examine the maps they generate to determine which model seems the most reliable between BR-NPA and the CNN, using the Score-CAM algorithm to explain the CNN's decision.
By reliable, we mean a model which uses biologically relevant features in the image.

\begin{figure}[tbh!p]  
    \centering
    \begin{subfigure}[t]{0.66\textwidth}
        \includegraphics[width=\textwidth]{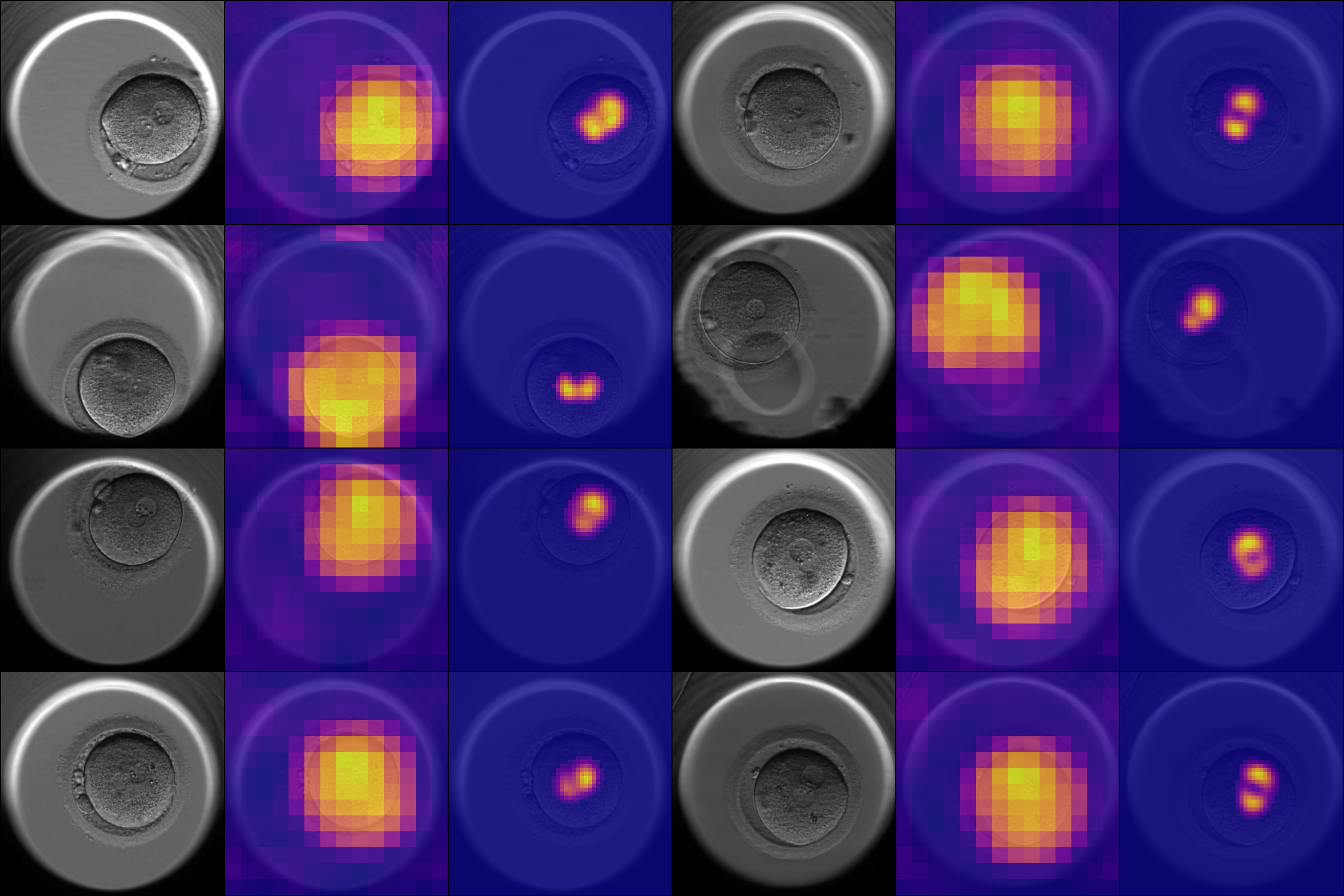} \caption{\label{tpna}}
    \end{subfigure}   
    \begin{subfigure}[t]{0.66\textwidth}
        \includegraphics[width=\textwidth]{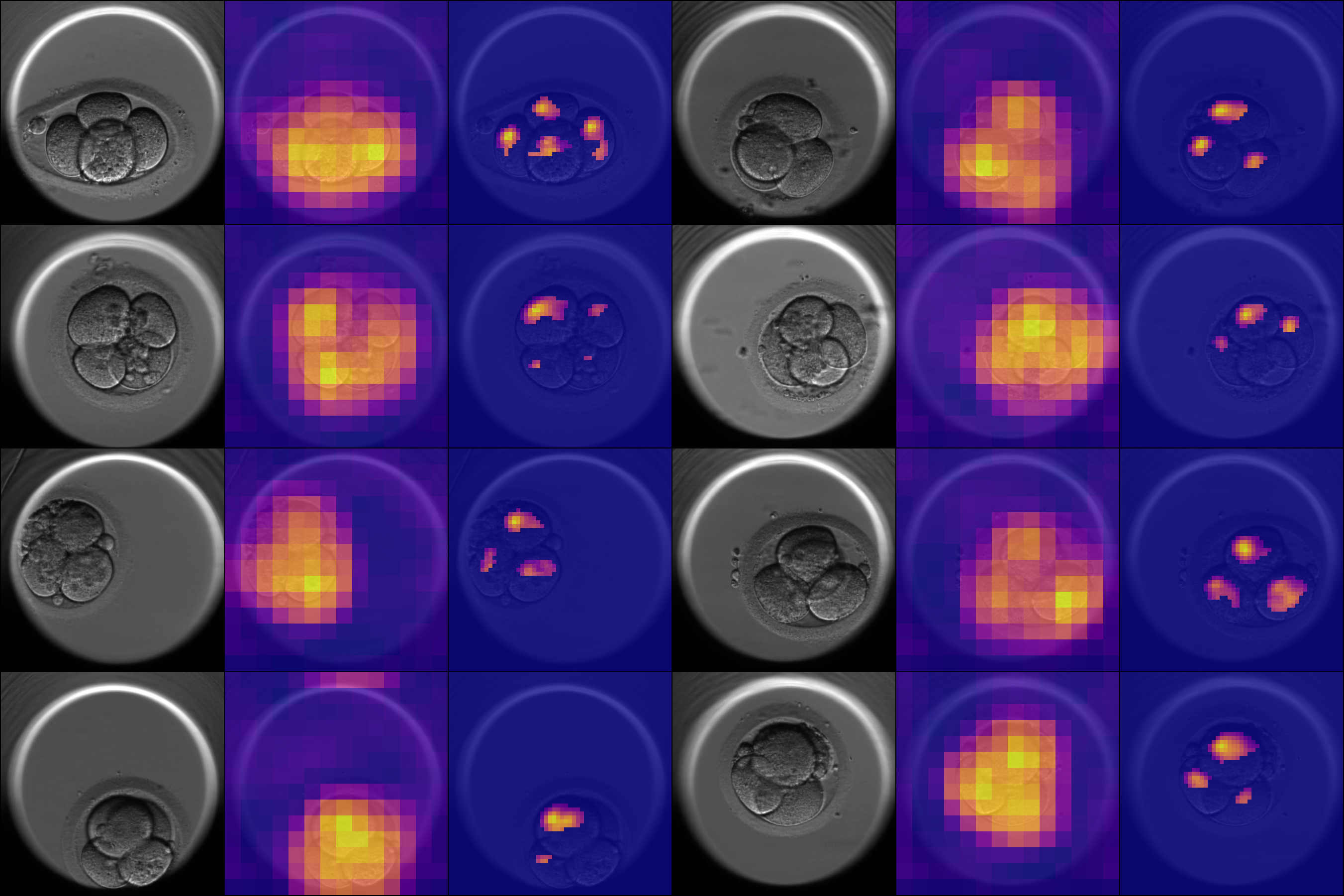} \caption{\label{t4}}
    \end{subfigure}  
    \begin{subfigure}[t]{0.66\textwidth}
        \includegraphics[width=\textwidth]{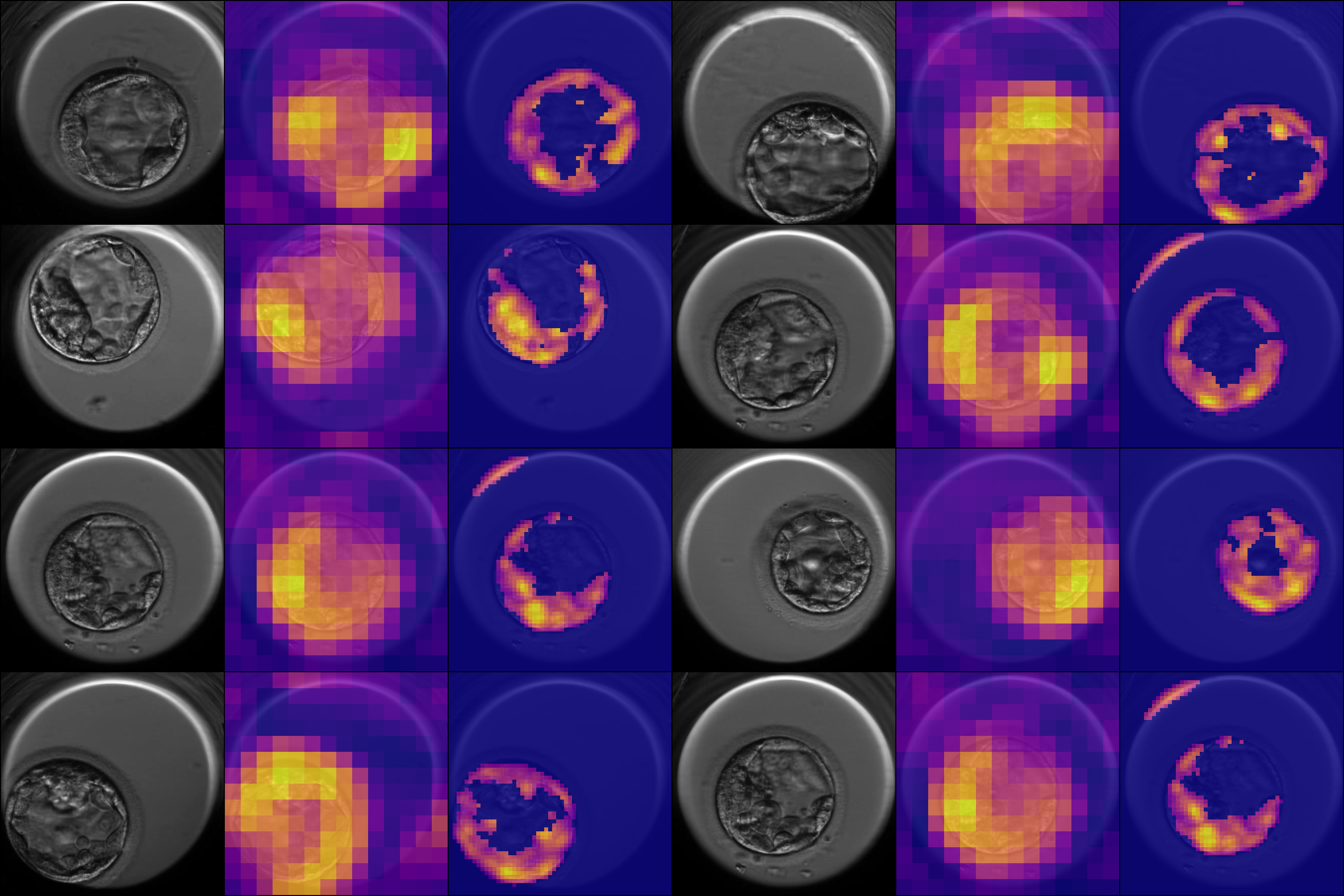} \caption{\label{tB}}
    \end{subfigure}  
    \caption{The saliency maps generated by the Score-CAM method and BR-NPA model (respectively middle and right columns) on stages sPNa (a), s4 (b), and sB (c).\label{attMaps}}
\end{figure}

\Cref{attMaps} shows examples of saliency maps generated by BR-NPA and Score-CAM during 3 stages: sPNa (\cref{tpna}), s4 (\cref{t4}) and sB (\cref{tB}).
In \cref{tpna}, BR-NPA focuses distinctly on the Pro-Nuclei (PN) (the nuclei of a sperm or an egg cell during the process of fertilization) during the sPNa stage whereas the focus of the CNN seems to be at an upper level as it highlights the whole embryo.
The focus of BR-NPA's attention on the PN is biologically relevant because the PN are only visible during this stage.
In \cref{tB} one can see that BR-NPA focuses on the wall of the embryo during the sB phase.
This stage marks the beginning of a new cell structure in the embryo where, instead of being arranged as a pack, cells start to specialize and notably some cells form the wall of the embryo.
Therefore, focusing on the embryo's wall is also biologically relevant as it is a marker of this phase.
On the other hand, Score-CAM produces low-resolution maps that can not highlight small details and focus on the embryo as a whole, a less biologically relevant level of focus given that it is visible during all stages.
Globally, BR-NPA's saliency maps are sharp and precise and highlight biologically relevant elements of the image.
However, this does not imply that BR-NPA focuses on more biologically relevant features than the CNN.
Indeed, the blurriness of Score-CAM's saliency maps is probably due at least to the low resolution of the CNN's feature maps ($14\times14$) compared to the resolution of BR-NPA's which is higher ($56\times56$).
This shows the interest in using high-resolution attention maps like BR-NPA maps to disambiguate the model's focus.

\section{Discussion}

Bastings et al. \cite{elephant} argue that post-hoc methods should be privileged over attention models when it comes to faithfulness, as post-hoc methods take the whole computation path into account whereas attention maps only reflect input importance at one point in the computation.
However, we showed here that depending on the way the map is evaluated (Highlight or Mask), attention models can provide superior faithfulness.

Also, the intuitions behind the metrics discussed here should be examined.
For example, the IIC metric suggests that the class score should increase once the salient parts of the image have been highlighted.
However, it is unclear if such a phenomenon would happen in practice.
Indeed, filling the masked areas with black pixels generates out-of-distribution samples that can induce an unexpected behavior of the model \cite{metricPaper}.
Moreover, ICC suggests that the confidence is increased by the presence of salient areas and decreased by the presence of non-salient areas, but the non-salient areas could also play a neutral role and not affect the confidence much.

More generally, the major difficulty in designing these metrics resides in measuring the impact of an area of the input image reliably.
Measuring the impact of an area by masking it is difficult because the score variation depends on if the other areas of the image are masked or not. 
To reach a consensus on how this measure should be done, the community has first yet to understand how exactly the score variation depends on the masking of other areas.
Moreover, the metrics studied here were all developed to quantify faithfulness but in practice adopt two distinct behavior and seems to favor different methods: attention models for the Mask group and post-hoc methods for the Highlight group.
Therefore, there should also be future works on the definition of faithfulness and its relationship with the approach type (attention models or post-hoc methods).

Finally, the practical relevance of the faithfulness metrics should be evaluated in a user study.
To what extent the reliability of the maps can improve, for example, the acceptability of the decision by a user remains an open question.

\section{Conclusion}

In this paper, we compared the faithfulness of the saliency maps generated by 9 different approaches, including 4 attention models and 5 post-hoc explanation methods. 
We showed low overall agreement between the metrics proposed in the literature and in particular demonstrated the tendency of metrics of the Highlight group to favor the post-hoc methods whereas metrics of the Mask group favor instead attention models.
We then visualized the saliency maps generated by the two best approaches, namely Score-CAM and BR-NPA, and showed that the low resolution of Score-CAM limits the insights that can be obtained whereas BR-NPA's maps highlight biologically relevant features.
Finally, we discussed the difficulty of measuring the impact of each image part reliably and future work.

\bibliographystyle{splncs04}
\bibliography{main}

\end{document}